\documentclass[sigplan,twocolumn,nonacm]{acmart}
\renewcommand\footnotetextcopyrightpermission[1]{}
\usepackage[english]{babel}
\usepackage{blindtext}
\usepackage{subcaption}
\usepackage{multirow}
\usepackage{enumitem}
\usepackage{needspace}
\newenvironment{denseitemize}{
    \begin{itemize}[topsep=2pt, partopsep=0pt, leftmargin=1.5em]
        \setlength{\itemsep}{3pt}
        \setlength{\parskip}{0pt}
        \setlength{\parsep}{0pt}
    }{\end{itemize}}
\AtBeginDocument{%
  }

\acmConference[Conference acronym 'XX]{Make sure to enter the correct
  conference title from your rights confirmation email}{June 03--05,
  2018}{Woodstock, NY}
\newcommand{\para}[1]{\noindent\textbf{#1}}
\newcommand{\sys}{\textsc{PEARL}\xspace}

\begin{document}

%%
%% The "title" command has an optional parameter,
%% allowing the author to define a "short title" to be used in page headers.
\title{\sys{}: Adaptive Prefill–Decode Execution with Elasticity for Agentic Reinforcement Learning}

%
% The "author" command and its associated commands are used to define
% the authors and their affiliations.
% Of note is the shared affiliation of the first two authors, and the
% "authornote" and "authornotemark" commands
% used to denote shared contribution to the research.
\author{Jiaan Zhu}
\authornote{Both authors contributed equally to this research.}
\email{zja\_pb17151780@mail.ustc.edu.cn}
\affiliation{%
  \institution{USTC}
  % \city{Hefei}
  % \state{Anhui}
  \country{China}
}
\author{Wei Gao}
\authornotemark[1]
\email{csgaowei@ust.hk}
\affiliation{%
  \institution{HKUST}
  % \state{Hongkong}
  \country{China}
}
\author{Youhui Bai}
\correspondingauthor
\email{youhuibai@ustc.edu.cn}
\affiliation{%
  \institution{USTC}
  % \city{Hefei}
  % \state{Anhui}
  \country{China}
}
\author{Zewen Jin}
\email{zevin@ustc.edu.cn}
\affiliation{%
  \institution{USTC}
  % \city{Hefei}
  % \state{Anhui}
  \country{China}
}
\author{Ju Huang}
\email{huangju.hj@alibaba-inc.com}
\affiliation{%
  \institution{Alibaba Group}
  % \city{Hefei}
  % \state{Anhui}
  \country{China}
}
\author{Siran Yang}
\email{siran.ysr@alibaba-inc.com}
\affiliation{%
  \institution{Alibaba Group}
  % \city{Hefei}
  % \state{Anhui}
  \country{China}
}
\author{Jiamang Wang}
\email{jiamang.wang@alibaba-inc.com}
\affiliation{%
  \institution{Alibaba Group}
  % \city{Hefei}
  % \state{Anhui}
  \country{China}
}
\author{Lin Qu}
\email{xide.ql@taobao.com}
\affiliation{%
  \institution{Alibaba Group}
  % \city{Hefei}
  % \state{Anhui}
  \country{China}
}
\author{Cheng Li}
\email{chengli7@ustc.edu.cn}
\affiliation{%
  \institution{USTC}
  % \city{Hefei}
  % \state{Anhui}
  \country{China}
}

% \orcid{1234-5678-9012}
% \author{G.K.M. Tobin}
% \correspondingauthor
% \authornotemark[1]
% \email{webmaster@marysville-ohio.com}
% \affiliation{%
%   \institution{Institute for Clarity in Documentation}
%   \city{Dublin}
%   \state{Ohio}
%   \country{USA}
% }
% \author{Anonymous Authors}

%%
%% By default, the full list of authors will be used in the page
%% headers. Often, this list is too long, and will overlap
%% other information printed in the page headers. This command allows
%% the author to define a more concise list
%% of authors' names for this purpose.
% \renewcommand{\shortauthors}{Trovato et al.}

%%
%% The abstract is a short summary of the work to be presented in the
%% article.

%%
%% The code below is generated by the tool at http://dl.acm.org/ccs.cfm.
%% Please copy and paste the code instead of the example below.
%%
\settopmatter{printfolios=true,printacmref=false}

\begin{abstract}

Multi-turn rollout dominates the cost of agentic reinforcement learning (RL). Asynchronous execution and elastic GPU resources can accelerate this stage, but adding rollout replicas yields diminishing returns while training GPUs remain idle between updates. We observe that effective resource use also depends on the prefill--decode (PD) configuration. Both the choice between colocation and disaggregation and the optimal PD ratio vary with the workload, making resource scaling and PD configuration interdependent. Exploiting this opportunity requires selecting effective configurations and realizing their benefits within transient resource-availability windows despite reconfiguration costs.

We present \sys{}, an asynchronous agentic RL system that coordinates external resource elasticity, temporary reuse of idle training GPUs, and adaptive PD execution. \sys{} maintains a unified GPU--worker--role state and uses runtime profiles to predict rollout batch completion time, accounting for environment-induced reductions in decode concurrency. It selects the PD mode and ratio under the current GPU budget and translates each decision into an incremental transition plan that minimizes worker and role changes. Cost-aware switching and borrowing policies suppress transitions with insufficient expected benefit while ensuring timely return of training GPUs.
Our evaluation show that \sys{} achieves $2.17$--$2.79\times$ the throughput of fixed-resource ROLL across different LLMs. Compared with RLBoost+, throughput improves by up to approximately 26.9\% for Qwen3-8B and 36.3\% for Qwen3-30B-A3B.

\end{abstract}

\maketitle
\pagestyle{plain}
\thispagestyle{plain}

\section{Introduction}
\label{sec:intro}

Agentic reinforcement learning (RL) equips large language models (LLMs) with capabilities such as coding, web navigation, and tool use through interaction with an environment~\cite{kimik2,gemini2.5}.
Its workflow comprises two stages, \texttt{rollout}, which generates interaction trajectories, and \texttt{training}, which updates the model using these trajectories.
Rollout proceeds over multiple turns, repeatedly alternating between response generation and environment feedback~\cite{ragen,kimik1.5}.
Repeated inference over growing contexts and sequential environment interactions make rollout the dominant cost in this workflow~\cite{kimik1.5,gao2026roserolloutservinggpus,rlboost}.

To alleviate this bottleneck, recent RL systems increasingly adopt asynchronous execution, overlapping rollout and training on separate GPU pools~\cite{roll,areal}. This approach is also used in industrial model training~\cite{glm5,mimov26,intellect2}. 
However, overlap alone does not eliminate the throughput imbalance between the two stages. Systems therefore exploit elastic resources, including preemptible cloud instances and spare capacity in shared clusters, to expand the rollout pool when additional GPUs become available~\cite{rlboost,deepseekv4,tensorhub,gao2026roserolloutservinggpus}.

Existing systems nevertheless struggle to translate this additional capacity into efficient rollout execution (see \S\ref{sec:motivation-existing-limits}). Our characterization of RLBoost+, which extends RLBoost~\cite{rlboost} to support asynchronous agentic RL, reveals sublinear scaling. 
Increasing rollout resources from 16 to 24 GPUs adds 50\% more GPUs but reduces rollout time by only 10.3\% (Figure~\ref{fig:rollout_train_duration}). 
This is because rollout typically scales to additional GPUs through data parallelism~\cite{dean2012large}, which partitions a fixed trajectory batch across more workers while replicating model weights. Each worker therefore processes fewer trajectories while incurring similar weight-access costs, reducing GPU efficiency.
Rollout still takes $4.5$--$5.7\times$ as long as training, leaving training GPUs idle after each update while they await the next trajectory batch. 
These idle windows expose another source of rollout capacity. We refer to external resource changes as \texttt{inter-scale elasticity} and temporary reuse of idle training GPUs for asynchronous RL as \texttt{intra-scale elasticity}.

Agentic rollout extends LLM inference with environment interactions, and each inference call consists of two phases, prefill and decode. Our measurements further show that effective resource use depends on how these two phases are executed.
First, the preferred PD execution mode depends on the workload. Under PD colocation, recurrent prefills in multi-turn rollout interfere with decoding even with chunked prefill~\cite{chunked-prefill} and prefix caching~\cite{prefix-cache} enabled (Figure~\ref{fig:pd_interference}). PD disaggregation isolates the two phases on separate GPU sets, but introduces KVCache transfers and partitions the available capacity.
Second, when disaggregation is preferable, the optimal PD ratio also varies with the workload, requiring adaptive allocation of prefill and decode workers.
Because resource arrivals and reclamations change the workload per worker and feasible allocations, both the PD mode and ratio must adapt to the current workload and GPU budget.

Coordinating resource elasticity with PD execution presents two challenges. 
First, the system must select a configuration that shortens rollout batch completion time under changing workloads and resources. 
GPU count alone is insufficient to predict this outcome because performance also depends on PD interference, effective decode concurrency, KVCache transfers, and pauses for environment interactions. 
Second, the system must realize the selected configuration within transient resource-availability windows. Reconfiguration involves worker initialization, weight synchronization, and handling unfinished trajectories, while borrowed training GPUs must be returned before the next update. 
These transition costs can outweigh the expected acceleration. Configuration selection must therefore account for both execution performance and the cost of switching, while respecting mandatory resource returns.

To this end, we present \sys{}, an asynchronous agentic RL system that jointly manages resource elasticity and adaptive PD execution. 
\sys{} represents both external scaling and training-GPU reuse through a unified GPU--worker--role state. Its PD Decision Engine uses runtime profiles to predict rollout batch completion time, accounting for environment-induced reductions in decode concurrency, and selects the execution mode and PD ratio under the current GPU budget. 
Switching thresholds suppress changes with insufficient expected benefit. A unified orchestration layer converts each target configuration into an incremental transition plan that minimizes worker and role changes while preserving unaffected workers. 
\sys{} initializes added workers in the background, synchronizes their weights before admission, and retains partial trajectories when workers leave. For training-GPU reuse, it retains worker processes across handoffs and activates borrowing only when the predicted benefit exceeds the activation and return costs.

We implement \sys{} atop ROLL~\cite{roll} and evaluate it with Qwen3-8B and Qwen3-30B-A3B~\cite{qwen3} on SWE-bench~\cite{swe}. 
Experiments replay real preemptible-GPU availability traces on up to 32 GPUs, with 16 reserved for training, four reserved for rollout, and up to 12 additional GPUs available to rollout. 
Training uses FSDP2, and rollout workers use tensor parallelism of degree two. Across the evaluated traces, \sys{} achieves $2.17$--$2.37\times$ the throughput of fixed-resource ROLL for Qwen3-8B and $2.43$--$2.79\times$ for Qwen3-30B-A3B. Compared with RLBoost+ under the same resource-availability traces, \sys{} improves throughput by up to approximately 26.9\% and 36.3\%, respectively.

\section{Background}
\label{sec:back}
\subsection{Agentic RL and Asynchronous Training}

\begin{figure}[!t]
\centering
\includegraphics[width=0.48\textwidth]{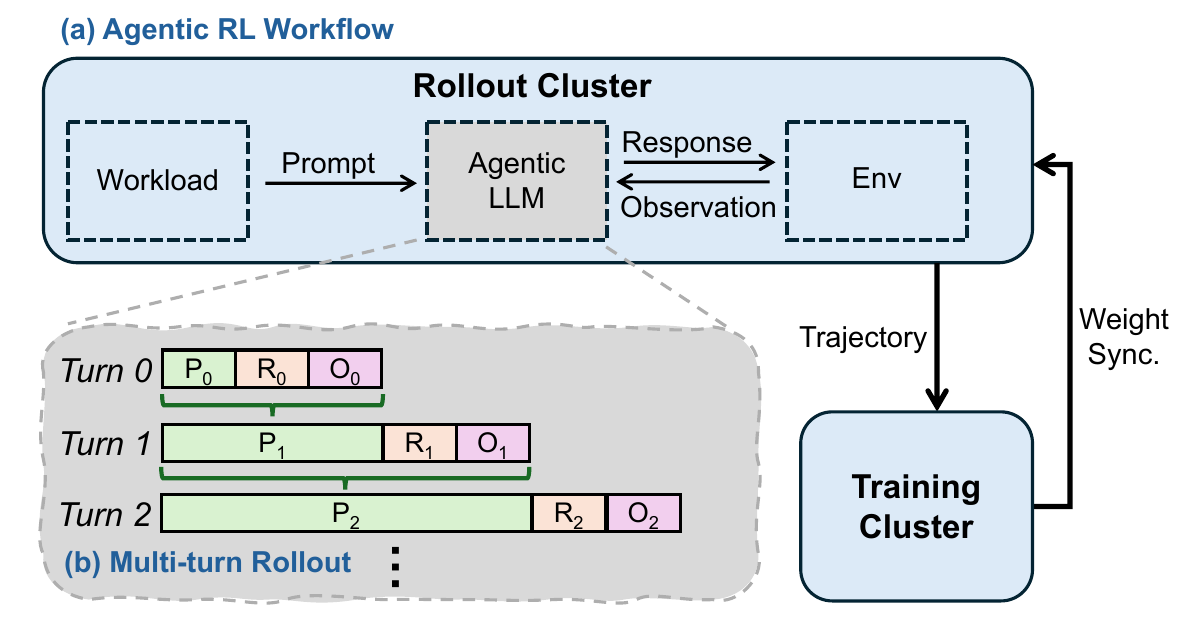}
% \vspace{-10pt}
\caption{Agentic RL workflow and multi-turn rollout.}
\label{fig:single_vs_multi}
% \vspace{-20pt}
\end{figure}

Agentic reinforcement learning (RL) trains large language models (LLMs) to solve tasks through repeated interaction with an environment~\cite{kimik2, gemini2.5, ragen}. By optimizing policies with rewards from these interactions, it develops capabilities such as planning, tool use, and adapting actions to feedback, with applications in software engineering~\cite{hao2025exploring, wu2025agentic, swe, swe-smith} and computer use~\cite{liu2025infigui, lu2026ui, luo2025gui}. 
As shown in Figure~\ref{fig:single_vs_multi}a, a typical agentic RL workflow comprises two stages, rollout and training. Rollout executes the policy to generate samples, while training consumes these samples to update model weights. The updated weights are then transferred back to rollout for the next step of workflow.

Figure~\ref{fig:single_vs_multi}b zooms in on the rollout stage, which combines LLM inference with multi-turn environment interaction. A task from the training dataset supplies the initial prompt $p_0$. \texttt{Prefill} processes it, and autoregressive \texttt{decoding} produces response $r_0$. The environment executes the action and returns observation $o_0$, yielding $p_1 = p_0 \oplus r_0 \oplus o_0$ for the next turn. Interaction continues until task completion, forming a \texttt{trajectory} that training uses to update model weights.

Multi-turn interaction makes rollout particularly expensive. 
Responses and observations accumulate across turns, expanding the context processed by subsequent inference calls, while autoregressive generation and environment execution introduce sequential dependencies within each trajectory. 
Together, these costs make rollout the dominant stage in our evaluated workload. 
% In our measurements on SWE-bench using 16 GPUs with colocated deployment in ROLL, rollout accounts for 91.3\% and 83.9\% of the end-to-end agentic RL execution time for Qwen-8B and Qwen3-30B-A3B, respectively.
For example, in our Qwen3-8B experiment on SWE-bench (Figure~\ref{fig:rollout_train_duration}), rollout still accounts for 81.8\% of the end-to-end agentic RL execution time, even when allocated four times as many GPUs as training.

To improve workflow efficiency, frameworks like AReaL~\cite{areal} and ROLL~\cite{roll} support asynchronous RL, an approach increasingly adopted in industrial LLM post-training~\cite{glm5, mimov26, intellect2}. Rollout and training run on separate GPU pools and overlap by allowing rollout to use weights with bounded staleness. This overlap reduces waiting from strict dependencies, but the throughput imbalance remains: when rollout is slower than training, training GPUs idle after each update while waiting for the next trajectory batch. Rollout efficiency therefore remains critical.

\subsection{Resource Elasticity in Agentic RL}

% Because rollout remains a major bottleneck, adding rollout resources can improve training throughput. However, the benefit of expansion diminishes as the resource pool grows: additional GPUs need not yield proportional throughput gains, while resource costs continue to rise. 
% This has motivated industrial systems to explore more elastic and cost-aware resource-management mechanisms. 
% DeepSeek and ByteDance Seed incorporate preemptible GPUs into their RL workflows, while Kimi and Prime Intellect support dynamic scaling of rollout-worker pools as resources become available or are reclaimed~\cite{deepseekv4, tensorhub, kimik1.5, intellect2}.
% Academic systems have similarly leveraged preemptible GPUs to accelerate rollout; for example, RLBoost and ROSE use such resources to increase rollout capacity at lower cost~\cite{rlboost, gao2026roserolloutservinggpus}.
% Together, these approaches allow the rollout resource pool to change with external resource availability, requiring agentic RL systems to resize the rollout cluster as resources are added or withdrawn.

Beyond asynchronous execution, rollout can exploit elastic GPUs. Public cloud providers offer discounted preemptible instances that add capacity when available but may be reclaimed~\cite{aws_ec2_spot_instances, google_cloud_gpus_on_spot_vms, bamboo}. Production clusters similarly share GPUs across priorities: DeepSeek-V4 uses a preemptible rollout service~\cite{deepseekv4}, Seed provisions workers on spot GPUs~\cite{tensorhub}, and Kimi and Prime Intellect scale rollout onto idle nodes~\cite{kimik1.5, intellect2}. Academic systems such as RLBoost~\cite{rlboost} and ROSE~\cite{gao2026roserolloutservinggpus} harvest preemptible and serving slack. In these settings, the rollout cluster must resize as resources come and go. 

\section{Motivation}
\label{sec:motivation}

% ------------------------------------------------------------
%  Sub-section: Dynamics in Asynchronous Rollout
% ------------------------------------------------------------
\begin{figure*}[!t]
    \centering
    \begin{subfigure}[b]{0.33\textwidth}
        \includegraphics[width=\linewidth]{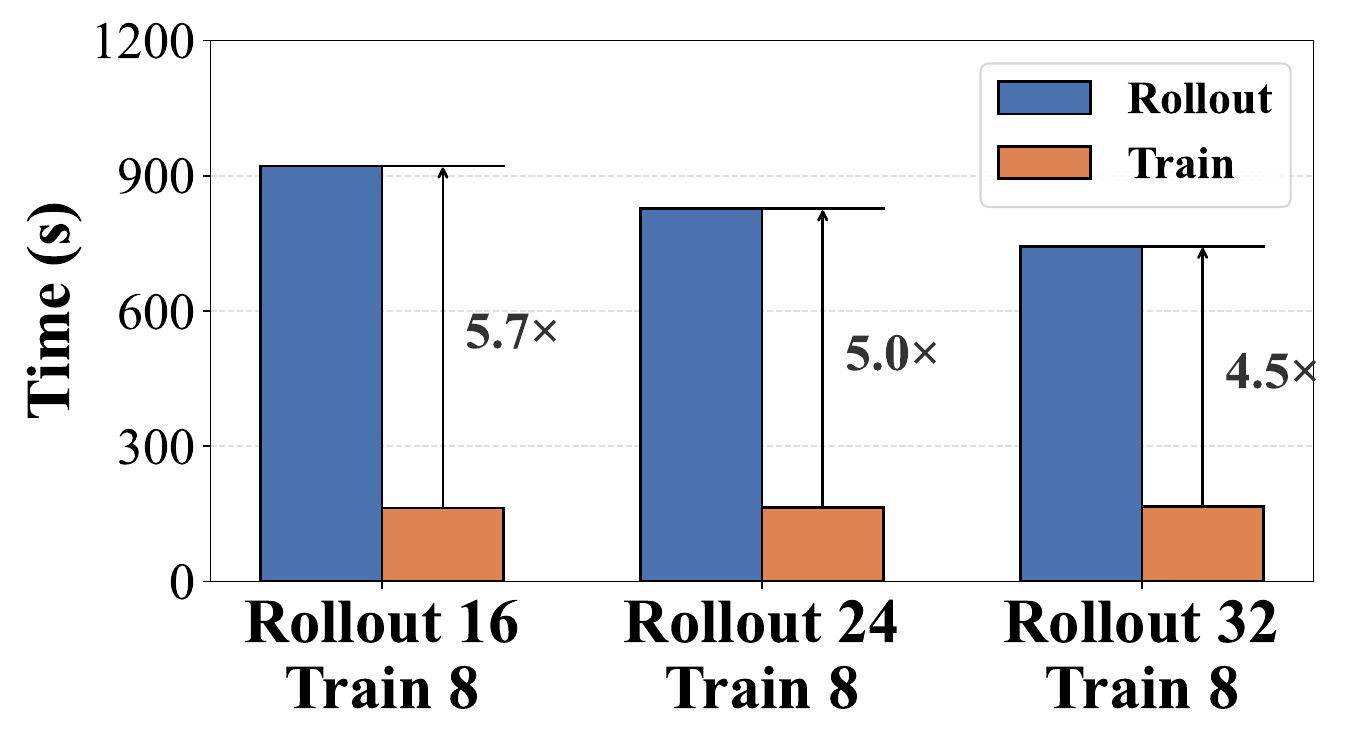}
        % \caption{Rollout and training durations for Qwen3-8B in the SWE environment as rollout resources increase from 16 to 32 GPUs, with training fixed at 8 GPUs. Rollout becomes faster with more GPUs, but the improvement is sublinear, and rollout remains longer than training.}
        \caption{Rollout and training durations.}
        \label{fig:rollout_train_duration}
    \end{subfigure}
    \hfill
    \begin{subfigure}[b]{0.33\textwidth}
        \includegraphics[width=\linewidth]{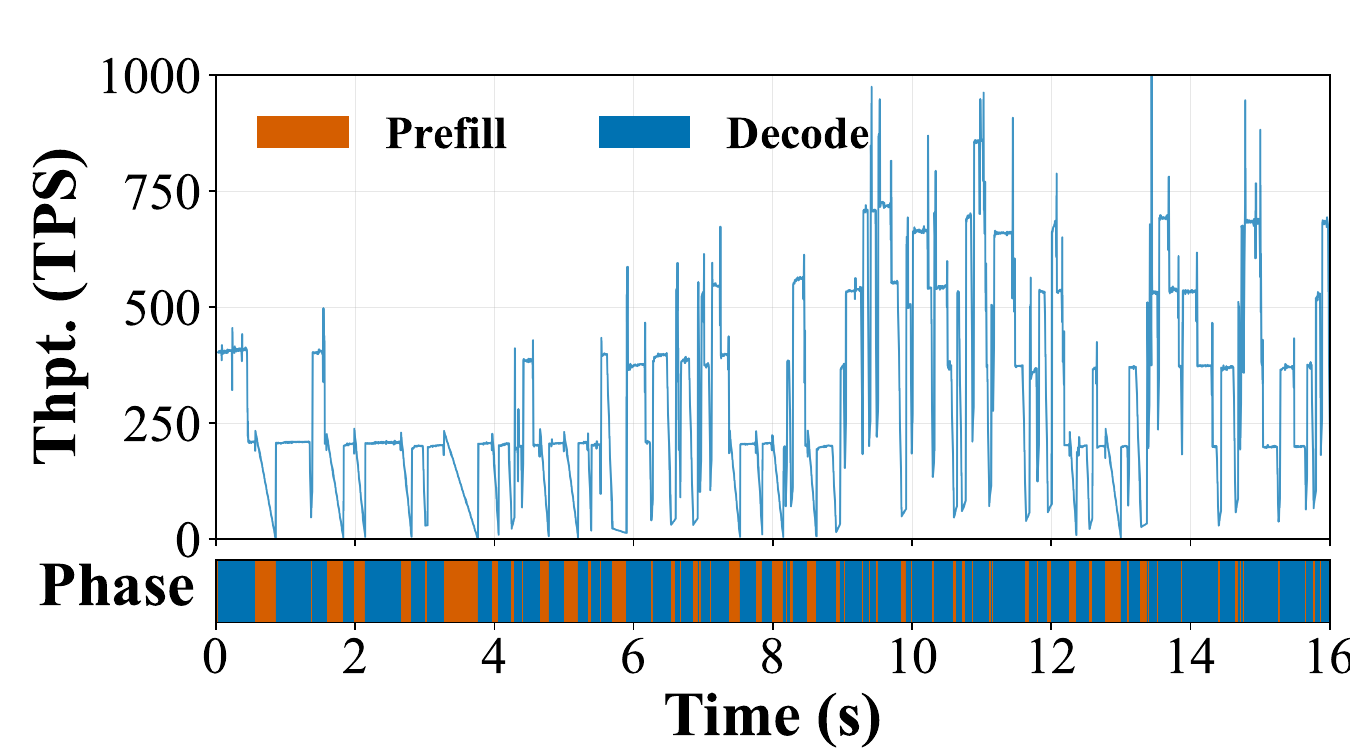}
        % \caption{Generation throughput of Qwen3-30B-A3 on SWE-bench during multi-turn rollout. Decode throughput is repeatedly interrupted by prefill insertions despite prefix caching.}
        \caption{Throughput during multi-turn rollout.}
        \label{fig:pd_interference}
    \end{subfigure}
    \hfill
    \begin{subfigure}[b]{0.33\textwidth}
        \includegraphics[width=\linewidth]{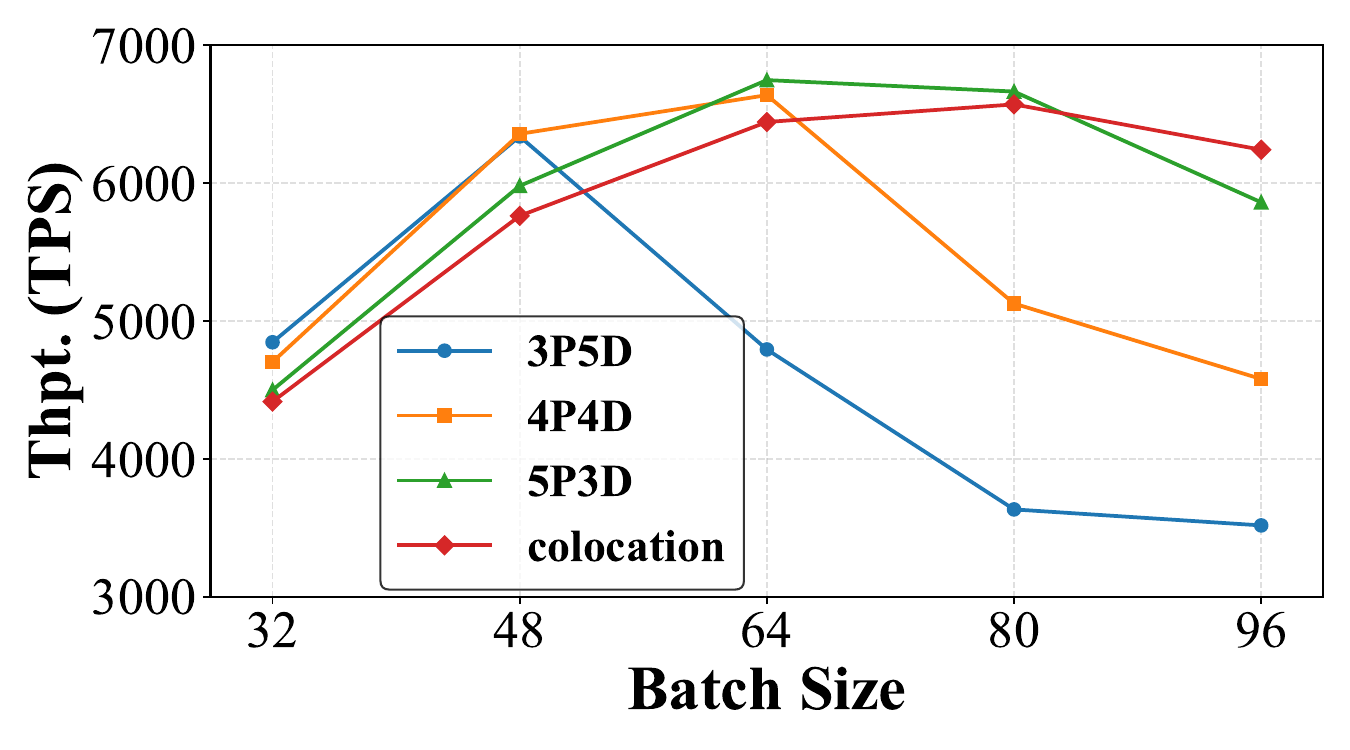}
        % \caption{Rollout throughput across batch sizes and PD ratios for Qwen3-30B-A3. The optimal PD ratio varies with batch size, and PD disaggregation does not always outperform PD colocation.}
        \caption{Throughput across BS and PD ratios.}
        \label{fig:throughput_vs_batch_size}
    \end{subfigure}
    % \vspace{-6pt}
    \caption{Computational characteristics of agentic RL workflows.(a) Rollout and training durations for Qwen3-8B as rollout resources increase from 16 to 32 GPUs, with training fixed at 8 GPUs.(b) Generation throughput of Qwen3-30B-A3B during multi-turn rollout. (c) Rollout throughput across batch sizes and PD ratios for Qwen3-30B-A3B.}
    \label{fig:motivation_all}
    % \vspace{-4pt}
\end{figure*}

\subsection{Underutilization of Elastic Resources}
\label{sec:motivation-dynamics}

To assess whether existing systems effectively exploit elastic resources, we characterize RLBoost+, our extension of RLBoost~\cite{rlboost} that supports agentic RL and asynchronous training, as described in the Section~\ref{sec:eval-setup}. 
We run Qwen3-8B~\cite{qwen3} on SWE-bench~\cite{swe} with 8 training GPUs and scale rollout from 16 to 32 GPUs through data parallelism (DP). The trajectory batch is partitioned across rollout workers, each hosting a replica of the model weights. The total trajectory batch size and per-worker model parallelism remain unchanged.
Figure~\ref{fig:rollout_train_duration} reveals resource underutilization, concerning additional external GPUs and the GPUs already allocated.

\para{Limited efficiency of external scale-out.} 
As shown in Figure~\ref{fig:rollout_train_duration}, provisioning additional GPUs reduces rollout time and consequently improves rollout throughput. 
However, the marginal benefit diminishes as the GPU allocation increases. Scaling from 16 to 24 GPUs increases the allocated resources by 50\% but reduces rollout latency by only 10.3\%. Further scaling to 32 GPUs yields an additional latency reduction of merely 8.9\%, with both reductions normalized to the rollout time on 16 GPUs. 
Therefore, the performance gains fall substantially short of proportional scaling, because data parallelism partitions the trajectory batch but replicates the model weights across workers. 
Each worker processes fewer trajectories while still reading the same model weights during decoding, amortizing weight-access costs over a smaller batch.
To assess whether scale-out alone can eliminate the bottleneck, consider an idealized scenario with unlimited rollout GPUs. Each DP worker processes one trajectory concurrently and per-worker model parallelism remains unchanged. 
Using the measured mean trajectory duration as an optimistic estimate, rollout still takes 216.3\,s, compared with 163.8\,s for training. This indicates that continuously adding rollout resources increases resource consumption without eliminating the rollout bottleneck.

\para{Idle capacity in the training pool.} The same imbalance also leaves training resources underutilized. Figure~\ref{fig:rollout_train_duration} shows that rollout takes $4.5$--$5.7\times$ as long as training. 
Although asynchronous execution overlaps both stages, training GPUs become idle after completing an update and wait for enough new trajectories to form the next batch. 
Under the idealized scale-out estimate above, rollout still takes approximately $1.3\times$ as long as training, leaving an estimated $24.3\%$ of training-GPU time idle in this workflow. 
These recurring idle windows provide additional capacity that can temporarily accelerate rollout before the GPUs return to training.

We refer to changes in rollout capacity obtained from outside the job's reserved GPU pool as \texttt{inter-scale} elasticity, and to temporary reuse of idle training GPUs within that pool as \texttt{intra-scale} elasticity. Both determine the resources available to rollout and must be considered jointly.

\subsection{Efficiency Degradation under PD Colocation}
\label{sec:motivation-pd-colocation}

We further investigate why rollout remains expensive despite additional GPU resources. Existing RL frameworks, including VeRL~\cite{verl}, ROLL~\cite{roll}, and slime~\cite{slime}, use LLM serving engines for rollout generation. 
A common deployment colocates prefill and decode on the same GPUs and adopts standard serving optimizations. 
Chunked prefill~\cite{chunked-prefill} divides long prefills into smaller chunks and batches them with decode requests, reducing decode stalls and peak activation memory usage. 
Prefix caching~\cite{prefix-cache} reuses KV states for previously processed prefixes, avoiding redundant computation across requests and interaction turns.

However, our measurements show that prefill still repeatedly interrupts decoding in agentic rollout, even with these optimizations enabled. 
This is because each environment interaction introduces new observations that require another prefill before generation resumes. 
Frequent prefill arrivals therefore continue to delay decoding and reduce generation throughput.
Figure~\ref{fig:pd_interference} illustrates this behavior for Qwen3-30B-A3B~\cite{qwen3} on SWE-bench. Decode throughput stays near 400~tokens/s per GPU between insertions but drops on prefill arrival. In this interval, prefill processes 96.68\% of tokens in 26.48\% of the elapsed time, while decode produces only 3.32\% of tokens but 73.52\% of the time. Decoding thus dominates execution time, yet prefill repeatedly disrupts it.

Consequently, adding GPUs under the same colocated strategy does not eliminate prefill--decode interference within each worker. 
This interference limits the effective use of resources from both \texttt{inter-scale} expansion and \texttt{intra-scale} reuse. 
Exploiting these resources therefore requires addressing PD interference alongside resource allocation.

% ------------------------------------------------------------
%  Sub-section: Opportunity of PD Disaggregation
% ------------------------------------------------------------

\subsection{Opportunity of PD Disaggregation}
\label{sec:motivation-pd-opportunity}
PD disaggregation places prefill and decode on separate GPU sets to eliminate interference~\cite{distserve}: prefill workers process prompts and transfer the KV cache to decode workers, allowing both phases to execute concurrently.

To assess its benefit for rollout of agentic RL, we extend ROLL~\cite{roll} with disaggregated execution and compare throughput for Qwen3-30B-A3B on SWE-bench across batch sizes, using colocation and different PD ratios (Figure~\ref{fig:throughput_vs_batch_size}). All use 16 GPUs as 8 workers with tensor-parallelism~\cite{shoeybi2019megatron} degree 2.

First, the preferred execution mode depends on the workload. The best disaggregated configuration outperforms colocation by 9.7\% at BS32, but the advantage narrows to 1.4\% at BS80 and reverses to a 6.1\% colocation win at BS96. Disaggregation trades reduced interference for KV-transfer overhead and a partitioned GPU pool; when transfer overhead dominates or either pool bottlenecks, colocation can be preferable.

Second, the optimal PD ratio depends on the workload. No single ratio is best across batch sizes: the best changes from \texttt{4P4D} at BS48 to \texttt{5P3D} at BS64. Shifting GPUs to prefill increases its concurrency, but under a fixed budget it leaves fewer decode GPUs, potentially making decode the bottleneck. A static ratio is therefore insufficient.

These observations connect PD configuration to resource elasticity: added GPUs change per-worker load and feasible allocations, so the execution mode and PD ratio must be re-selected from the current workload and GPU budget.

% ------------------------------------------------------------
%  Sub-section: Limitations of Existing Solutions
% ------------------------------------------------------------

\begin{table}[t]
  \centering
  \caption{Comparison of rollout deployment adaptation and PD configurations in representative RL systems. Resource adaptation denotes inter-scale elasticity and intra-scale training-GPU reuse; workload adaptation denotes deployment changes in response to workload.}
  % \vspace{-4pt}
  \label{tab:solution-comparison}
  \footnotesize
  \setlength{\tabcolsep}{1.5pt}
  \resizebox{\columnwidth}{!}{%
  \begin{tabular}{@{}c c c c c c@{}}
    \toprule
    % Account for the extra vertical space introduced by the cmidrules.
    \multirow{2}{*}[-\dimexpr(\aboverulesep+\belowrulesep+\cmidrulewidth)/2\relax]{\textbf{Category}} & \multirow{2}{*}[-\dimexpr(\aboverulesep+\belowrulesep+\cmidrulewidth)/2\relax]{\textbf{System}} & \multicolumn{2}{c}{\textbf{Adaptation}} & \multicolumn{2}{c}{\textbf{PD Configuration}} \\
    \cmidrule(lr){3-4}\cmidrule(lr){5-6}
    & & \textbf{Resource} & \textbf{Workload} & \textbf{Mode} & \textbf{P/D Ratio} \\
    \midrule
    \multirow{3}{*}{\shortstack{Fixed \&\\Async RL}}
    & AReaL\,\cite{areal} & \ding{55} & \ding{55} & Colocated & -- \\
    & ROLL\,\cite{roll} & \ding{55} & \ding{55} & Colocated & -- \\
    & Laminar\,\cite{laminar} & \ding{55} & \ding{55} & Colocated & -- \\
    \midrule
    \multirow{4}{*}{\shortstack{Elastic\\RL}}
    & RLBoost\,\cite{rlboost} & Inter-scale & \ding{51} & Colocated & -- \\
    & ROSE\,\cite{gao2026roserolloutservinggpus} & Inter-scale & \ding{51} & Colocated & -- \\
    & SeamlessFlow\,\cite{seamlessflow} & Intra-scale & \ding{55} & Colocated & -- \\
    & BiDiRL\,\cite{bidirl} & Intra-scale & \ding{55} & Colocated & -- \\
    \midrule
    \multirow{3}{*}{\shortstack{PD-\\Disaggregated\\RL}}
    & RollArt\,\cite{rollart} & \ding{55} & \ding{55} & Coloc.\ or Disag. & Static \\
    & NeMo RL\,\cite{nemorl_pd} & \ding{55} & \ding{55} & Coloc.\ or Disag. & Static \\
    & Slime\,\cite{slime} & \ding{55} & \ding{55} & Coloc.\ or Disag. & Static \\
    \midrule
    & \sys & \textbf{Both} & \ding{51} & \textbf{Coloc.\,$\leftrightarrow$\,Disag.} & Dynamic \\
    \bottomrule
  \end{tabular}%
  }
  % \vspace{-15pt}
\end{table}

\subsection{Limitations of Existing Solutions}
\label{sec:motivation-existing-limits}

\begingroup
\setlength{\emergencystretch}{3em}

The preceding observations show that exploiting elastic resources requires adapting both rollout capacity and PD execution.
The preferred execution mode and P/D ratio depend on the workload and available resources.
Table~\ref{tab:solution-comparison} summarizes how representative RL systems address these requirements.

\para{Fixed and asynchronous RL systems.}
AReaL~\cite{areal}, ROLL~\cite{roll}, and Laminar~\cite{laminar} improve training efficiency by overlapping rollout and policy updates.
Their rollout deployments use fixed resource pools with colocated prefill and decode.
Asynchronous execution alone does not enable these deployments to absorb external GPU capacity or reuse idle training GPUs, leaving both forms of resource elasticity unexploited.

\para{Elastic RL systems.}
RLBoost~\cite{rlboost} and ROSE~\cite{gao2026roserolloutservinggpus} provide \texttt{inter-scale} elasticity through external preemptible GPUs and spare serving capacity, respectively.
SeamlessFlow~\cite{seamlessflow} and BiDiRL~\cite{bidirl} provide \texttt{intra-scale} elasticity by reusing training resources for rollout.
These systems demonstrate complementary ways to obtain additional capacity, but retain colocated PD execution.
As \S\ref{sec:motivation-pd-opportunity} shows, the preferred execution mode changes with workload.
Adding rollout capacity alone therefore does not resolve whether prefill and decode should remain colocated or how resources should be divided between them.

\para{PD-disaggregated RL systems.}
RollArt~\cite{rollart}, NeMo RL~\cite{nemorl_pd}, and Slime~\cite{slime} support PD-disaggregated rollout, but use statically configured execution modes and P/D ratios.
A configuration chosen at launch need not remain efficient when external GPUs arrive or are reclaimed, or when training GPUs become available for rollout and are later returned.
Supporting PD disaggregation alone does not provide adaptation to the changing resource budget and workload.

\para{Why not simply combine existing pieces?}
Combining elasticity with PD disaggregation requires coordinated decisions and transitions.
Resource changes alter both the feasible P/D allocations and the workload per worker, so configuration selection must track the current resource state.
Applying a choice requires handling in-flight requests, synchronizing weights, and changing worker roles while meeting resource-reclamation and training-resumption requirements.
These transitions incur costs that can outweigh the gains from optional expansion or borrowing.
\sys{} therefore couples workload-aware configuration selection with cost-aware reconfiguration, jointly considering \texttt{inter-scale} and \texttt{intra-scale} resources to reduce rollout-batch makespan.
\par\endgroup

\newcommand{\Circled}[1]{\ding{\the\numexpr171+#1\relax}}
\section{Design}
\label{sec:design}
\subsection{System Overview}
\label{sec:design-overview}

\begin{figure}[!t]
\centering
\includegraphics[width=0.49\textwidth]{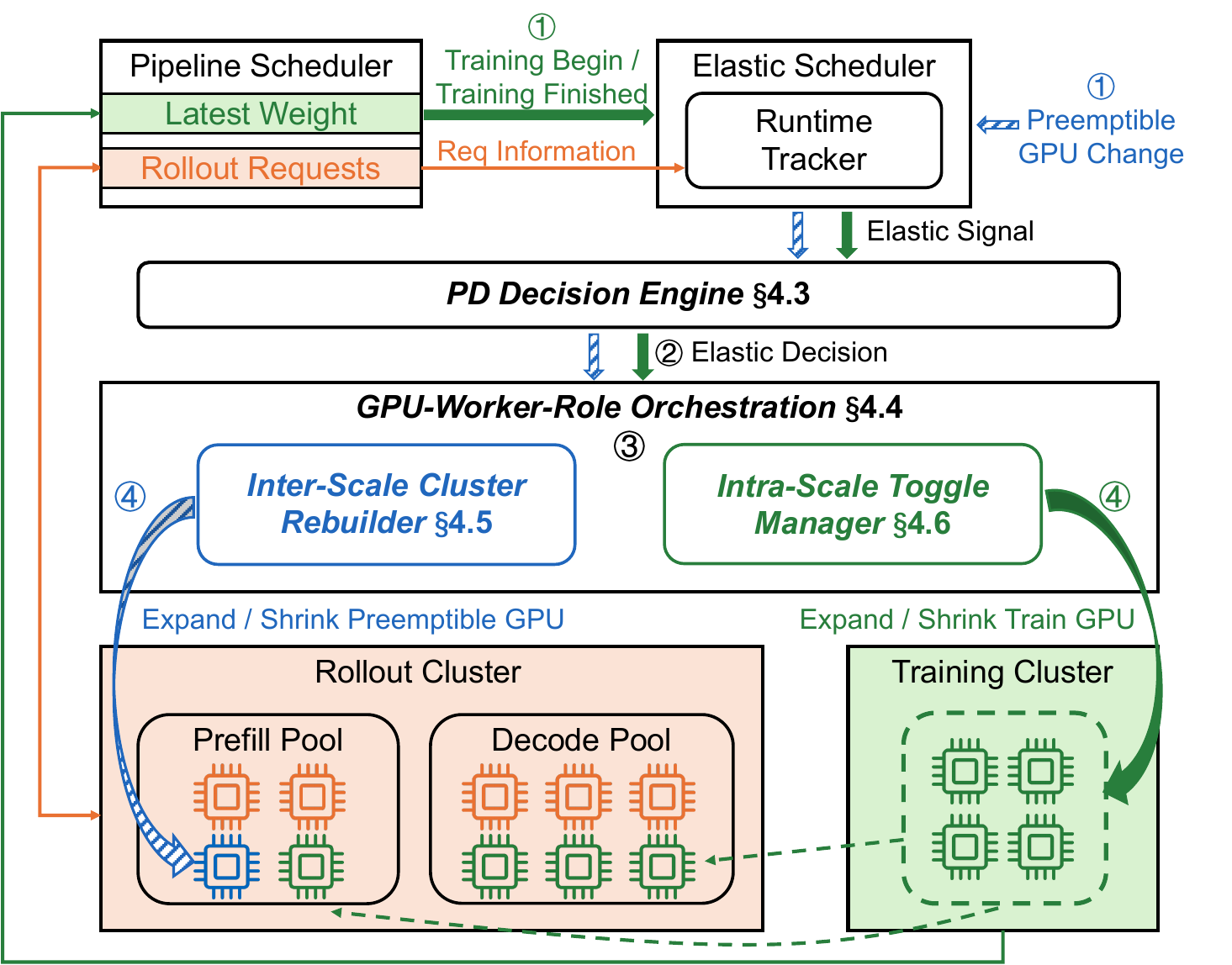}
\caption{Architecture of \sys{}.}
% \vspace{-8pt}
\label{fig:system_overview}
% \vspace{-12pt}
\end{figure}

\subsubsection{Architecture}
Figure~\ref{fig:system_overview} shows the architecture of \sys{}.
The Pipeline Scheduler runs the asynchronous RL hot path: its router dispatches rollout requests to the serving workers, and it collects trajectories and synchronizes weights.
The Elastic Scheduler reacts to Training Begin/Finished notifications and preemptible-GPU changes. Its Runtime Tracker maintains a workload profile from request telemetry.
The PD Decision Engine uses the profile and the current rollout GPU budget to choose an execution mode (colocation or disaggregation) and role allocation that minimizes predicted batch makespan.
GPU--Worker--Role Orchestration turns that allocation into an incremental reconfiguration plan, run by two executors: the Inter-Scale Cluster Rebuilder for preemptible-GPU changes and the Intra-Scale Toggle Manager for training-GPU reuse.

The rollout cluster generates trajectories with prefill, decode, and colocated workers, while the training cluster keeps fixed workers and GPU bindings. Training GPUs can be borrowed for rollout during idle windows (\S\ref{sec:design-intra-scale}).
The remainder of this section formalizes the optimization, then details the decision engine, orchestration, and the two executors.

\subsubsection{Workflow}
All elastic events, including training notifications and preemptible-GPU changes, traverse the same four stages. The Elastic Scheduler updates the rollout budget \Circled{1}. The PD Decision Engine selects the execution mode and role allocation \Circled{2}. GPU--Worker--Role Orchestration produces an incremental reconfiguration plan \Circled{3}. The responsible module executes it \Circled{4}. The two triggers differ only in how the budget changes and in the handoff constraints.

\noindent\textbf{A complete training step.}
During rollout, the Pipeline Scheduler dispatches requests while the Runtime Tracker profiles the workload. Training needs a complete batch, so training GPUs sit idle until the remaining trajectories finish. The Toggle Manager can borrow them during this window (\S\ref{sec:design-intra-scale}).
Once the batch is ready, Training Begin triggers a shrink: borrowed training GPUs are reclaimed, leaving only dedicated rollout GPUs \Circled{1}, the PD Decision Engine re-selects mode and roles \Circled{2}, Orchestration schedules their return first \Circled{3}, and the Intra-Scale Toggle Manager drains their requests, retains partial trajectories, and releases the GPUs \Circled{4}. Dedicated rollout workers keep serving.

After training updates the model and the Pipeline Scheduler syncs new weights, Training Finished triggers a grow: the budget again includes all training GPUs \Circled{1}, the same decision and orchestration path produces a plan \Circled{2}--\Circled{3}, and the Intra-Scale Toggle Manager admits workers on borrowed training GPUs only if the benefit-aware gate permits and their runtime state and weight versions are ready for serving \Circled{4}. Otherwise rollout continues on dedicated workers alone.

\noindent\textbf{Changes in preemptible-GPU availability.}
Preemptible-GPU changes use the same stages \Circled{1} through \Circled{3}, and only execution differs. For a shrink, the Inter-Scale Cluster Rebuilder removes the reclaimed workers from routing, drains them as in the training-triggered shrink, and returns their GPUs. If this breaks the PD topology, the decision stage already picks a feasible replacement (\S\ref{sec:design-formalization}). For an expansion, it initializes workers in the background and admits them only after weight sync, so requests never hit a half-ready worker. Only affected workers change, and the rest keep serving.

Training-GPU borrowing and preemptible-GPU scaling differ: returning borrowed GPUs is mandatory because training cannot wait, while borrowing and expansion are optional and happen only when predicted gain exceeds handoff cost. Since both share one decision and execution path, the workload, not the GPU source, decides whether added resources become prefill, decode, or colocated workers.

\subsection{Problem Formalization}
\label{sec:design-formalization}

\noindent\textbf{Unified optimization view.}
We formalize the choice of execution mode and role allocation under a time-varying budget as one optimization problem. The system chooses whether to disaggregate prefill and decode or colocate them and, under disaggregation, each role's worker count.

\noindent\textbf{Objective.}
Because rollout is the bottleneck in our workloads, \sys{} improves training throughput by reducing trajectory-batch completion time rather than per-request latency. We therefore use the batch makespan $T_{\mathrm{roll}}(\pi)$, the completion time of the last of the $B$ trajectories under configuration $\pi$, as the online criterion, and select the feasible configuration $\pi^\star$ with the minimal predicted makespan, re-solving as the workload and resource budget change.

\noindent\textbf{Decision variables and feasible configurations.}
For rollout step $k$, let $N_k$ denote the number of currently available rollout workers. \texttt{Inter-scale} and \texttt{intra-scale} events update this budget, and the Runtime Tracker supplies the latest workload profile and platform parameters. The decision variables are $\pi=(N_P,N_D,N_C)$ with $N_P+N_D+N_C=N_k$, where $N_P$, $N_D$, and $N_C$ are the numbers of active prefill, decode, and colocated workers. The number of training workers $N_T$ is fixed, so the complete target configuration is $(N_T,N_P^\star,N_D^\star,N_C^\star)$. A configuration is either fully colocated ($N_P=N_D=0$) or fully disaggregated ($N_C=0$ with $N_P,N_D\geq 1$). The feasible set $\Pi'(N_k)$ contains the assignments satisfying the capacity constraints below. If a shrink would remove all prefill or all decode workers, the system either reassigns surviving workers or falls back to colocation.

\noindent\textbf{Capacity constraints.}
We obtain the feasible set $\Pi'(N_k)$ by enforcing two capacity constraints. The prefill workers must handle the token workload $\lambda_P = B \cdot S \cdot \Delta L$ per step without building queues that delay decode. Under disaggregation, the decode workers must also hold the entire batch's KV cache: $B \cdot K_{\mathrm{seq}} \leq N_D \cdot H_{\max}$, where $K_{\mathrm{seq}}$ is the KV-cache size per trajectory and $H_{\max}$ is the HBM capacity per worker.

\subsection{PD Decision Engine}
\label{sec:design-pd-engine}
The PD Decision Engine solves this problem by building closed-form makespan models for colocation and disaggregation from hardware specs, model architecture, and runtime telemetry, enumerating the feasible disaggregated configurations, and selecting the one with the minimal predicted batch makespan.

\subsubsection{Bubble-Aware Makespan Models}
Within a rollout step, each trajectory alternates between prefill insertions and decode, so prefill and decode requests interleave across the batch. They compete for the same GPUs under colocation but run on separate pools under disaggregation. We therefore model the two modes separately.

\noindent\textbf{PD colocation.}
Prefill and decode execute serially on shared GPUs.
We model the batch makespan as
\begin{equation}
    \begin{aligned}
        T_{\mathrm{coloc}}=S\Biggl[&
        \underbrace{\frac{B\cdot\Delta L}{N}\cdot t_{\mathrm{prefill}}}_{T_{\mathrm{prefill}}^{\mathrm{turn}}}
        +\underbrace{r\left(A+b\frac{B}{N}\right)}_{T_{\mathrm{decode}}^{\mathrm{turn}}}+T_{\mathrm{over}}\Biggr]+T_{\mathrm{env}}.
    \end{aligned}
    \label{eq:design-colocated-makespan}
\end{equation}
The batch has $B$ trajectories, $N=N_k$ rollout workers, and $S$ interaction turns on the slowest trajectory. Each turn adds $\Delta L$ prompt tokens and generates $r$ tokens. The per-token prefill time $t_{\mathrm{prefill}}$ is the computation per token divided by GPU compute capacity. In the decode term, $A$ is the fixed per-iteration cost, including weight reads and kernel dispatch, and $b$ is the marginal cost of each additional concurrent sequence, mainly from extra KV-cache HBM reads. $T_{\mathrm{over}}$ covers overheads such as switching between prefill and decode, and $T_{\mathrm{env}}$ is the environment interaction time.

The model fixes the decode batch size at $B/N$. It assumes work conservation: when trajectories leave decode for environment interactions, prefill consumes the freed GPU capacity. The model therefore captures these interactions through $S$ and $T_{\mathrm{env}}$ without per-turn concurrency correction.

\noindent\textbf{PD disaggregation.}
Under disaggregation, the $N_k$ rollout workers are split into $N_P$ prefill workers and $N_D$ decode workers, with $N_P+N_D=N_k$. Prefill and decode run on separate GPU pools and can execute concurrently. During environment interactions, trajectories sit in neither queue, so the effective decode batch size drops. Unlike colocation, disaggregated decode GPUs cannot reuse idle cycles for prefill, so environment interactions reduce decode throughput.

We quantify this effect by the average fraction of trajectories that are actively decoding, $f$, or equivalently the bubble ratio $\theta=(1-f)/f$. The effective decode concurrency then drops from $B$ to $B/(1+\theta)$. The decode-path batch makespan is
\begin{equation}
    T_{\mathrm{decode}}=S\cdot r\cdot
    \left(A+b\frac{B}{N_D(1+\theta)}\right)(1+\theta).
    \label{eq:design-disaggregated-decode}
\end{equation}
The prefill-path batch makespan is
\begin{equation}
    T_{\mathrm{prefill}}=S\left(
        \frac{B\cdot\Delta L}{N_P}\cdot t_{\mathrm{prefill}}
        +\frac{B\cdot\Delta L\cdot K}{\mathrm{BW}}
    \right).
    \label{eq:design-disaggregated-prefill}
\end{equation}
$t_{\mathrm{prefill}}$ is the same as in the colocated model, $K$ is the KV transfer volume per token, and $\mathrm{BW}$ is the available bandwidth between the two pools. The second term lower-bounds the cross-pool KV-transfer time because it assumes ideal, uncontended use of the full bandwidth. The disaggregated makespan is the longer of the two paths plus environment interaction time, $T_{\mathrm{dis}}=\max(T_{\mathrm{prefill}},T_{\mathrm{decode}})+T_{\mathrm{env}}$.

\subsubsection{Online Decision and Switching Control}

After each rollout step, the Runtime Tracker refits the makespan-model parameters ($S$, $r$, $\Delta L$, $\theta$, $A$, $b$, and $t_{\mathrm{prefill}}$) from request telemetry. Upon each elastic signal, the PD Decision Engine enumerates feasible disaggregated configurations over $N_P\in[1,N_k-1]$ (with $N_D=N_k-N_P$), discards those that violate the capacity constraints in \S\ref{sec:design-formalization}, compares the best against colocation, and returns the target configuration. The orchestration layer then executes the required incremental changes (\S\ref{sec:design-orchestration}).

Applying a new configuration is not free: role rebinding, router updates, and CUDA graph recapture can take tens of seconds, so when the current and target configurations have similar predicted makespans, the transition can cost more than it saves. The PD Decision Engine therefore applies mode-dependent hysteresis: a switch is admitted only when the alternative configuration's predicted makespan is lower by more than a threshold, with a larger threshold for colocation--disaggregation mode switches than for ratio adjustments within the same mode. Hysteresis suppresses thrashing from performance noise, but does not block forced switches needed to keep a feasible role configuration when the current topology becomes infeasible.

\subsection{GPU--Worker--Role Orchestration}
\label{sec:design-orchestration}

The PD Decision Engine outputs a target configuration that specifies how many workers each role needs, but not which physical workers should take these roles or how to reach the target from the current configuration. GPU--Worker--Role Orchestration solves this placement and transition problem: it maintains a consistent GPU--worker--role mapping across elastic events while minimizing reconfiguration overhead.

The two kinds of elastic resources have different lifecycles: preemptible GPUs physically join and leave the cluster, while training GPUs stay in place and only change which phase executes on them. The Orchestration therefore partitions GPUs into two zones. The dedicated rollout zone holds the GPUs reserved for rollout and expands or shrinks as preemptible instances join and leave. The overlap zone holds the training GPUs, which rollout can borrow during idle windows. Rollout workers placed in this zone are \emph{overlap workers}. Training workers and their GPU bindings remain fixed: borrowing changes execution ownership, not the training topology. The GPUs available for rollout at any time are the union of the currently available GPUs in the two zones.

Elastic events in these zones pose two challenges: concurrent events can leave different system components with conflicting views of the same worker, and rebuilding each target configuration from scratch turns a local change into global overhead. The Orchestration addresses them with two mechanisms, a unified GPU--worker--role state and minimal-change plan generation.

\subsubsection{A Unified GPU--Worker--Role State}

Inter-scale events, intra-scale events, and role adjustments can all arrive during the same rollout. If the router, the executors, and the Pipeline Scheduler keep separate views, they may disagree on whether a worker is serving, transitioning, or due back to training. The result can be requests routed to departing workers or training and rollout contending for the same GPU memory. The Orchestration therefore represents resource state as a single unified GPU--worker--role mapping shared by all system components. The mapping has two links: the GPU-to-worker link binds each worker to a fixed set of physical GPUs, and the worker-to-role link records each worker's current role and activity status. This single source of truth links the PD Decision Engine's target configuration to concrete execution and prevents different system components from interpreting it independently. Logically, the state is a table keyed by stable worker identifiers $w$:
\begin{equation}
    \mathcal{M}=\{w\mapsto(G_w,R_w,a_w)\},
    \label{eq:design-worker-state}
\end{equation}
where $G_w$ is the GPU set bound to worker $w$, $R_w$ is its role (prefill, decode, colocated, or training), and $a_w$ is its membership status in the active set: \emph{active} workers currently receive and serve rollout requests (the set $\mathcal{W}$), \emph{inactive} workers keep their processes alive but serve no requests (e.g., an overlap worker whose GPUs are currently running training), and \emph{transitioning} workers are being added to or removed from the active set, or reassigned, and must drain or finish initialization before they can serve.

\noindent\textbf{Decoupling physical bindings from logical roles.}
The mapping keeps physical bindings stable and lets only logical roles change. Each worker occupies a stable slot in the table and binds to a fixed set of GPUs for its lifetime, so scaling never renumbers surviving workers and new workers simply reuse empty slots. Fixed bindings still allow time-sharing: in the overlap zone, a training worker and an overlap rollout worker share the same $G_w$, and the role $R_w$ and activity $a_w$ determine which phase currently uses the GPUs. Rollout workers can change roles or leave and rejoin the active set without altering their bindings. The three forms of dynamics therefore reduce to two logical updates on this state: changing the active worker set (membership) or reassigning roles among surviving workers. Mapping updates are serialized so concurrent events never apply to stale state, while reconfiguration operations execute asynchronously.

\subsubsection{Minimal-Change Plan Generation}

Rebuilding each target configuration from scratch recreates workers that need not change. Worse, a rebuilt cluster renumbers the surviving workers, and communication groups, request routing, and weight broadcasts all reference workers by index, so every survivor must rejoin these structures. A change that touches a few workers then becomes a cluster-wide reinitialization. Placements with identical role counts are therefore not equivalent. The orchestration layer instead turns the target configuration into a plan that preserves existing bindings and roles wherever possible. Stable slots separate membership changes from role changes among surviving workers. For current and target active sets $\mathcal{W}$ and $\mathcal{W}'$, the workers in $\mathcal{W}'\setminus\mathcal{W}$ must join and those in $\mathcal{W}\setminus\mathcal{W}'$ must leave. For each role $r$, let $n_{\mathrm{src}}(r)$ and $n_{\mathrm{dst}}(r)$ be its counts in the surviving set $\mathcal{S}=\mathcal{W}\cap\mathcal{W}'$ and in the target configuration. At most $\min(n_{\mathrm{src}}(r),n_{\mathrm{dst}}(r))$ survivors can keep role $r$, and the excess must switch. The change count therefore decomposes into membership and role changes:
\begin{align}
    D_{\mathrm{worker}}&=|\mathcal{W}'\setminus\mathcal{W}|+|\mathcal{W}\setminus\mathcal{W}'|,
    \label{eq:design-membership-changes}\\
    D_{\mathrm{role}}&=\sum_r\Bigl[n_{\mathrm{src}}(r)-\min\bigl(n_{\mathrm{src}}(r),n_{\mathrm{dst}}(r)\bigr)\Bigr].
    \label{eq:design-role-changes}
\end{align}
Because the two worker sets are disjoint, no placement can trade a membership change for a role change: every join and every departure is fixed by the target configuration, and among the survivors the excess of each role must switch. Keeping each existing role up to its target count and filling the remaining deficits with added workers and reassigned survivors therefore achieves exactly the minimum change count $D_{\mathrm{worker}}+D_{\mathrm{role}}$.

The orchestration layer emits a concrete reconfiguration plan $(\mathcal{W}_{\mathrm{add}},\mathcal{W}_{\mathrm{remove}},\rho)$ to the execution modules, where $\mathcal{W}_{\mathrm{add}}$ and $\mathcal{W}_{\mathrm{remove}}$ capture membership changes and $\rho$ maps surviving workers to new roles. Because both the Inter-Scale Cluster Rebuilder and the Intra-Scale Toggle Manager consume the same plan format, the two GPU sources share a single execution path; they differ only in the urgency of returning GPUs.

% ------------------------------------------------------------
\subsection{Inter-Scale Cluster Rebuilder}
\label{sec:design-inter-scale}

The Rebuilder handles changes in the dedicated rollout zone caused by preemptible-GPU arrivals and reclamations, executing the add/remove portions of the orchestration plan while rollout continues on unaffected workers.

Reconfiguration is inherently asymmetric. Shrinking is cheap and urgent: reclaimed GPUs must be returned promptly, but interrupted workers may hold partial trajectories. Expansion is expensive and deferrable: new workers pay a large cold-start cost and cannot be admitted until fully initialized with the latest weights, yet rollout cannot pause to wait.

\noindent\textbf{Shrinking.}
The Rebuilder stops dispatching new requests to the affected workers, interrupts unfinished requests while retaining partial trajectories, and returns the GPUs synchronously. Subsequent execution resumes from these partial trajectories without interrupting rollout on other workers.

\noindent\textbf{Expansion.}
New workers initialize in the background while existing workers continue serving. Because training may update weights during this window, the Rebuilder uses \emph{lazy weight propagation}: it maintains the latest available weight snapshot and synchronizes new workers only after they initialize, avoiding both stale startup checkpoints and frequent checkpoint writes. Once all added workers have finished initialization, weight synchronization, and runtime preparation, the Rebuilder applies \emph{atomic finalization} to admit the entire group at once. This pauses admission of new requests briefly but does not block requests already executing. Finally, an \emph{all-or-nothing admission} policy cancels the whole expansion if any worker fails to initialize, preventing a partial topology from being exposed to the router.

% ------------------------------------------------------------
\subsection{Intra-Scale Toggle Manager}
\label{sec:design-intra-scale}

The Manager borrows training GPUs during idle windows between training steps, time-sharing existing training GPUs with rollout rather than adding physical GPUs.

Idle compute alone does not ready a GPU for rollout because training and rollout share the same memory budget; pausing training does not free rollout memory. Repeated process destruction at every handoff would incur cold-start costs that erase the benefit within short windows.

\noindent\textbf{Process Retention and Device-State Handoff.}
The Toggle Manager decouples process lifetimes from device-state residency. It retains training and rollout processes across handoffs and swaps only the device state needed for the active phase: training releases GPU memory when borrowing begins, and overlap workers drain before returning. Returning is mandatory because training cannot proceed until borrowed GPUs are released. Rollout on dedicated GPUs continues uninterrupted.

\noindent\textbf{Benefit-aware activation.}
Process retention reduces repeated initialization but not activation and return costs. The Toggle Manager therefore activates borrowing only when the predicted benefit exceeds the handoff cost. It compares $\mathrm{ded}$, using only dedicated rollout resources, with $\mathrm{full}$, using all training GPUs as overlap workers. With $R$ of the $B$ trajectories not yet completed, it enables borrowing when
\begin{equation}
    \frac{R}{B}\left(T_{\mathrm{ded}}-T_{\mathrm{full}}\right) > T_{\mathrm{toggle}},
    \label{eq:design-toggle-gate}
\end{equation}
where $T_{\mathrm{toggle}}$ is the incremental cost of activation and return. This cost-aware gate prevents the Toggle Manager from paying handoff overhead for borrowing windows that are too short to amortize it.

\needspace{6\baselineskip}
\section{Implementation}
\label{sec:implementation}

\sys{} extends ROLL with elastic rollout management, weight synchronization, and runtime workload tracking, using SGLang's~\cite{sglang} native PD disaggregation for rollout workers. For intra-scale toggle, SGLang's integrated torch memory saver releases resident weights, KVCache, and CUDA graphs~\cite{distserve, mooncake} when training begins while retaining the rollout process; these states are restored when GPUs become available. Under PD disaggregation, restoration also refreshes communication metadata for KV buffers. Interrupted requests preserve their generated-token prefix and interaction progress in the environment, then resubmit the prefix to an available worker to resume generation.

\para{Training workers.}
\sys{} extends ROLL's Megatron~\cite{shoeybi2019megatron} and FSDP~\cite{fsdp} backends with complete weight snapshots and topology-aware weight updates. Snapshots reuse weights aggregated during normal updates, avoiding additional checkpoints. The controller stores only version, metadata, and references. New rollout workers asynchronously fetch the snapshot, load it, and hand it to the local inference process.

\para{Worker lifecycle.}
\sys{} manages training and rollout workers as Ray~\cite{ray} actors, tracking their GPU bindings, roles, and service states. External resource managers submit inter-scale changes through Ray, while the training pipeline triggers intra-scale borrowing and returns without recreating workers.

\para{Runtime statistics.}
For each step, \sys{} records interaction turns, newly added input tokens, generated tokens, response tokens retained in the next context, and environment-processing time. It aggregates these statistics across steps to update the PD decision engine's workload profile, exploiting similarity between adjacent steps without assuming a stationary workload. Exponential moving average~\cite{ema} for toggle-enabled and toggle-disabled steps estimate switching costs online and update the intra-scale gate.
% \needspace{6\baselineskip}
\section{Evaluation}
\label{sec:eval}
We evaluate \sys{} along five dimensions: end-to-end throughput under changing GPU availability, the performance impact of PD disaggregation and intra-scale toggle, the quality of the PD Decision Engine's ratio selection, the sensitivity to workload and reconfiguration overhead. 
\subsection{Experimental Setup}
\label{sec:eval-setup} 

\begin{figure}[!t]
\centering
\includegraphics[width=0.48\textwidth]{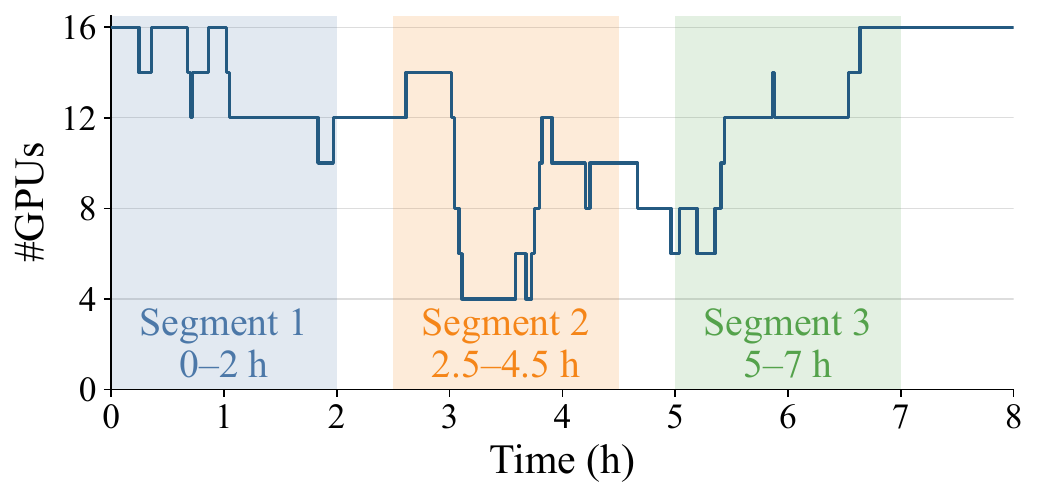}
\caption{Dedicated rollout GPU availability and the three two-hour segments used in our experiments. The count includes 4 reserved GPUs and up to 12 preemptible GPUs; the 16 reserved training GPUs are excluded.}
\label{fig:eval-traces}
\end{figure}

\para{Models and training configuration.}
We train Qwen3-8B and Qwen3-30B-A3B on software-engineering tasks from SWE-bench using GRPO with the FSDP2 backend.
Unless otherwise specified, we use a group size of 16, a rollout batch size of 64, a maximum trajectory length of 32K tokens, and a maximum policy staleness of 1 training step.
All rollout workers---prefill, decode, and colocated---use tensor parallelism of degree 2 (TP2).

\para{Cluster setup.}
We conduct our experiments on a cluster of 32 NVIDIA H800 GPUs, each with 80\,GB of GPU memory.
GPU nodes are interconnected via 800\,Gbps InfiniBand.
We allocate 16 reserved GPUs to training, providing sufficient memory for Qwen3-30B-A3B.
In the end-to-end experiments, \sys{} and RLBoost+ use 4 reserved rollout GPUs and up to 12 additional preemptible GPUs, whose availability follows the replayed trace below.

\para{Resource traces.}
Following RLBoost~\cite{rlboost}, we replay three representative two-hour segments from the real GPU availability trace provided by Bamboo~\cite{bamboo}.
Figure~\ref{fig:eval-traces} shows the selected segments:
Segment~1 exhibits frequent availability changes with abundant GPU resources;
Segment~2 exhibits frequent changes with limited resources;
and Segment~3 provides moderate resources with fewer changes.

% zewen:
\if0
\para{Cluster setup and resource traces.}\hspace{0pt}
We conduct our experiments on a cluster of 32 NVIDIA H800 GPUs, each with 80\,GB of GPU memory.
GPU nodes are interconnected via 800\,Gbps InfiniBand.
We replay real preemptible GPU availability trace from Bamboo~\cite{bamboo}, using representative two-hour segments following the methodology used by RLBoost~\cite{rlboost}.
Figure~\ref{fig:eval-traces} shows the three segments used in our experiments.
Segment~1 has frequent availability changes but generally maintains a large pool of available GPUs.
Segment~2 also changes frequently, but provides fewer available GPUs overall.
Segment~3 has fewer availability changes and a moderate number of available GPUs.
Together, the three segments cover frequent changes under both abundant and constrained capacity, as well as relatively stable intermediate capacity.
We allocate 16 reserved GPUs to training to provide sufficient memory for Qwen3-30B-A3B.
Based on the minimum GPU availability in the trace, we allocate 4 reserved GPUs to rollout to maintain a basic asynchronous deployment with separate training and rollout resources.
We then add up to 12 preemptible GPUs according to the trace, allowing the rollout GPU count to follow the replayed availability.
\fi

\para{Baselines.}
We compare \sys{} with three baselines in the end-to-end experiments, all using the same 16 reserved training GPUs:
\begin{denseitemize}
    \item \textbf{ROLL.} This asynchronous baseline uses a fixed allocation of 4 reserved rollout GPUs and no preemptible resources.
    \item \textbf{ROLL-Large.} This variant of ROLL uses a larger fixed pool of reserved rollout GPUs. For Segments~1--3, the mean rollout GPU counts are 13.6, 9.0, and 11.8. Rounding each up to the nearest multiple of 2 for TP2 gives fixed allocations of 14, 10, and 12 GPUs, respectively. These allocations avoid the preemption and instance-startup overheads associated with resource changes.
    \item \textbf{RLBoost+.} RLBoost+ is our adaptation of RLBoost~\cite{rlboost} for asynchronous agentic RL. It uses the same reserved and preemptible rollout resources and replays the same availability traces as \sys{}.
    % Author check: original RLBoost uses training-GPU seeding. Document how
    % RLBoost+ adapts this mechanism and which original optimizations it retains.
\end{denseitemize}

\para{Throughput metrics.}
For each step, system throughput is the sum of the token counts for rollout generation and training, divided by the step duration.
We report throughput in tokens per second, showing both per-step values over time and averages for each trace segment.
% Author check: specify segment averaging, the asynchronous step boundaries,
% treatment of interrupted/recomputed/unused tokens, and the measurement window
% for startup and reconfiguration. These details require the experiment logs.

% \begin{figure*}[!t]
%     \centering
%     \begin{subfigure}[b]{0.24\textwidth}
%         \includegraphics[width=\linewidth]{figures/rlboost_trace_12h_preemptible.pdf}
%         \caption{Rollout and training durations.}
%         \label{fig:rollout_train_duration}
%     \end{subfigure}
%     \hfill
%     \begin{subfigure}[b]{0.75\textwidth}
%         \includegraphics[width=\linewidth]{figures/e2e-avg.pdf}
%         \caption{Throughput during multi-turn rollout.}
%         \label{fig:pd_interference}
%     \end{subfigure}
%     \vspace{-6pt}
%     \caption{}
%     \label{fig:motivation_all}
%     \vspace{-4pt}
% \end{figure*}

\begin{figure*}[!t]
\centering
\includegraphics[width=0.96\textwidth]{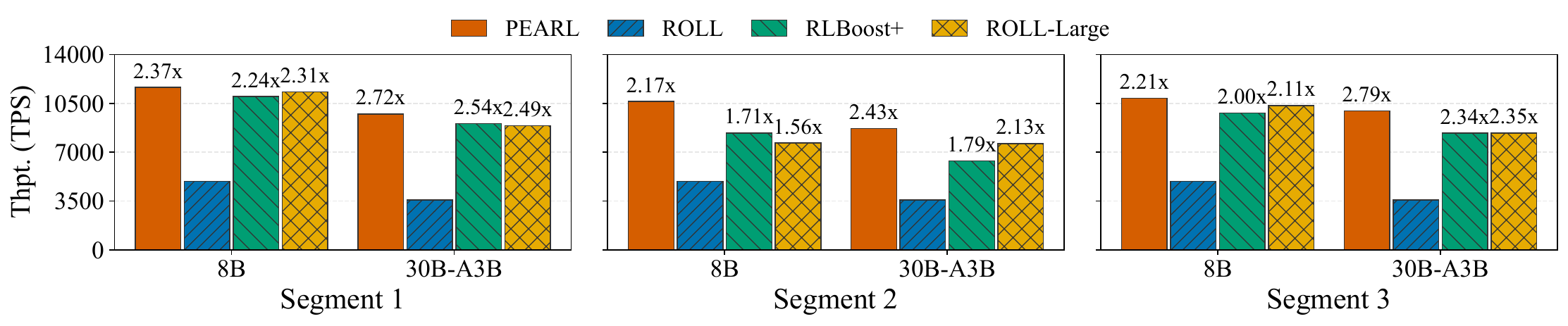}
\caption{Average system throughput for each model and trace segment. Bar annotations are normalized to ROLL. RLBoost+ shares \sys{}'s external rollout resource availability and training GPU allocation; ROLL-Large uses a fixed, larger reserved rollout pool.}
\label{fig:eval-throughput-average}
\end{figure*}

\Needspace{15\baselineskip}
\subsection{End-to-End Performance}
\label{sec:eval-e2e}

% Numerical source: the original Fig. 6 (e2e-avg.pdf), as requested by the author.
% Gains relative to RLBoost+ are approximate because its bar annotations are
% rounded to two decimals; each gain is (ROSTER / RLBoost+ - 1) * 100%.
\para{Overall throughput.}
\sys{} improves average throughput over RLBoost+ by approximately 5.8--26.9\% for Qwen3-8B and 7.4--36.3\% for Qwen3-30B-A3B across the three trace segments (Figure~\ref{fig:eval-throughput-average}).
Both systems face the same external rollout resource availability and use 16 reserved training GPUs, making RLBoost+ our primary comparison for execution efficiency under changing resources.
Relative to fixed-resource ROLL, \sys{} achieves $2.17$--$2.37\times$ and $2.43$--$2.79\times$ the throughput for the two models, respectively.
This latter comparison captures the combined benefit of additional preemptible resources and \sys{}'s execution mechanisms.

\para{Impact of resource availability.}
The advantage over RLBoost+ is largest in Segment~2, where the dedicated rollout pool averages 9.0 GPUs: throughput improves by approximately 26.9\% for Qwen3-8B and 36.3\% for Qwen3-30B-A3B.
In Segment~1, with 13.6 rollout GPUs on average, the improvements narrow to approximately 5.9\% and 7.4\%.
Segment~3 lies between these cases, averaging 11.8 rollout GPUs and yielding gains of approximately 10.6\% and 19.0\%.
The larger gains under constrained rollout capacity are consistent with the fixed-resource ablations in \S\ref{sec:eval-ablation}, which show greater benefit from reusing idle training GPUs when dedicated rollout capacity is limited.
The trace comparisons establish this pattern across the evaluated conditions; they do not isolate the contribution of each mechanism.

\para{Comparison with fixed provisioning.}
\sys{} is also competitive with ROLL-Large, which uses 14, 10, and 12 reserved rollout GPUs for Segments~1--3, respectively.
These fixed allocations slightly exceed each segment's average dedicated rollout capacity and avoid preemption and resource-change startup overheads.
\sys{} achieves higher measured average throughput for both models in all three segments, although its Qwen3-8B result in Segment~1 is close to the fixed deployment.
ROLL-Large therefore provides a stable-provisioning reference; its fixed allocation does not represent a performance upper bound or an identical resource schedule.

\begin{figure*}[!t]
\centering
\includegraphics[width=0.98\textwidth]{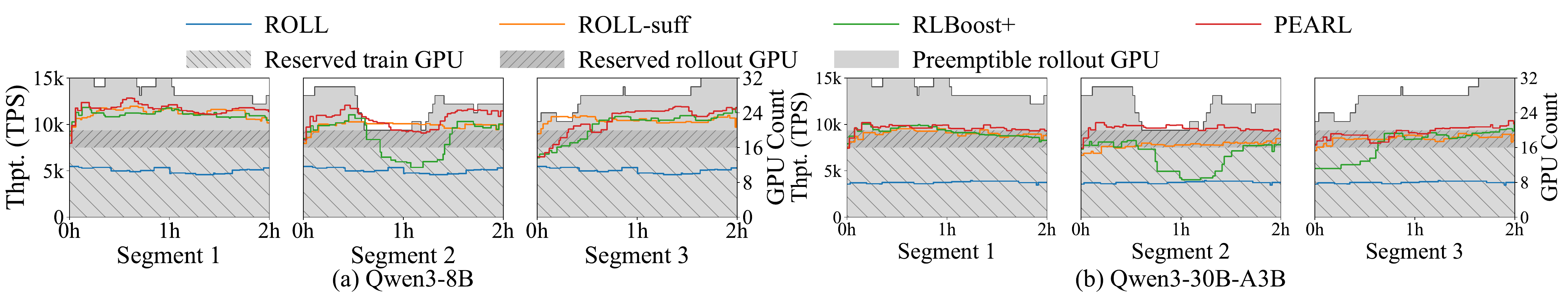}
\caption{Per-step system throughput over three segments. The stacked background shows the shared physical GPU budget: 16 reserved training GPUs, 4 reserved rollout GPUs, and up to 12 preemptible rollout GPUs. \sys{} and RLBoost+ can utilize all the GPUs, while ROLL uses the reserved allocation and ROLL-Large uses a fixed rollout pool of 14, 10, or 12 GPUs.}
\label{fig:eval-throughput-time}
\end{figure*}

\para{Behavior under resource changes.}
Figure~\ref{fig:eval-throughput-time} shows throughput evolution under the same trace segments.
During the pronounced resource reduction in the middle of Segment~2, RLBoost+'s per-step throughput drops more sharply than \sys{}'s for both models.
As resources return, RLBoost+'s throughput recovers and the gap narrows.
This interval illustrates where \sys{} sustains its advantage within the trace, complementing the segment averages.
Because throughput is measured per step, these curves do not resolve instantaneous reconfiguration latency.
The following ablation study examines the separate contributions of PD disaggregation, intra-scale toggle, and its benefit-aware gate.

\subsection{Ablation Study}
\label{sec:eval-ablation}

\begin{figure}[!t]
\centering
\includegraphics[width=0.48\textwidth]{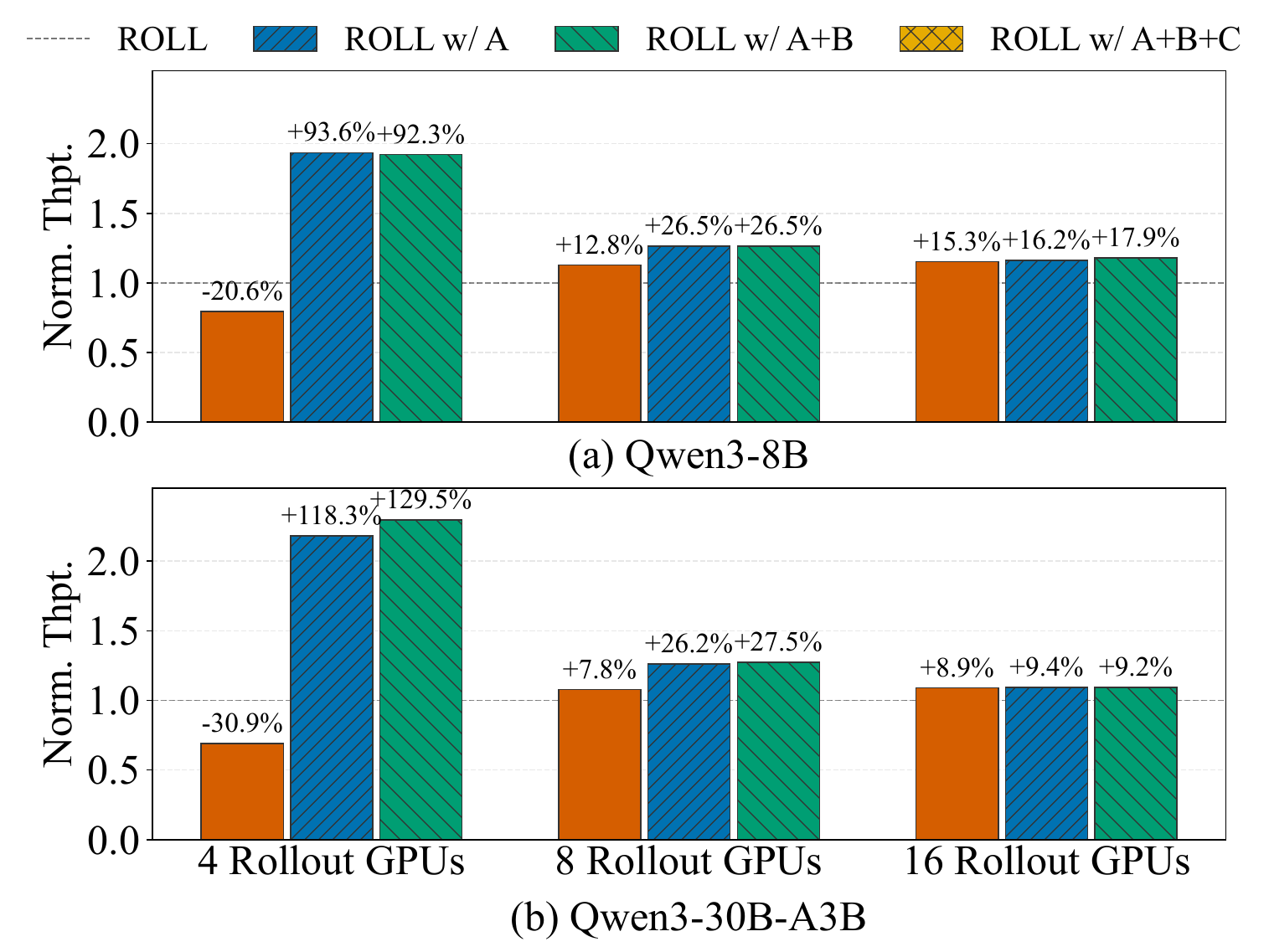}
\caption{Ablation study with 16 training GPUs and fixed allocations of 4--16 rollout GPUs. Throughput is normalized to ROLL. A: optimal PD-disaggregation ratio; B: ungated intra-scale toggle; C: benefit-aware intra-scale gate. \sys{} combines all three components.}
\label{fig:eval-ablation}
\end{figure}

To isolate the contributions of PD disaggregation, intra-scale toggle, and the intra-scale gate, we evaluate Qwen3-8B and Qwen3-30B-A3B under fixed resource allocations.
Each experiment uses 16 training GPUs and 4 to 16 dedicated rollout GPUs, with the same training parameters as the end-to-end experiments.
Figure~\ref{fig:eval-ablation} reports each configuration's throughput relative to the ROLL baseline.
We then add, in order, PD disaggregation with the optimal PD ratio, ungated intra-scale toggle, and the benefit-aware intra-scale gate, with the final configuration corresponding to \sys{}.

With 4 rollout GPUs, the PD Decision Engine would select PD colocation. For this ablation, however, we force the PD-disaggregated configuration so that the figure exposes the cost and benefit of PD disaggregation at this resource point. 
PD disaggregation reduces throughput by 20.6\% for Qwen3-8B and 30.9\% for Qwen3-30B-A3B.
This high compute pressure also makes intra-scale toggle highly effective: adding the ungated toggle raises throughput by 93.6\% and 118.3\%, respectively.
Adding the gate further increases the Qwen3-30B-A3B gain to +129.5\%.
For Qwen3-8B, the initial estimate of toggle overhead was too high, so the gate conservatively disabled toggle during several early phases that would have benefited from it; the gated configuration therefore achieves +92.3\%, slightly below the +93.6\% of ungated toggle. The Qwen3-30B-A3B result with 16 rollout GPUs shows the same effect.

Across 8--16 rollout GPUs, PD disaggregation improves throughput by 12.8--15.3\% for Qwen3-8B and 7.8--8.9\% for Qwen3-30B-A3B.
Adding ungated intra-scale toggle changes these gains to 16.2--26.5\% and 9.4--26.2\%, respectively.
With the intra-scale gate, the gains become 17.9--26.5\% for Qwen3-8B and 9.2--27.5\% for Qwen3-30B-A3B.

These results indicate that the mechanisms address different resource conditions.
When dedicated rollout capacity is limited, intra-scale toggle borrows idle training GPUs and offsets the high compute pressure of rollout.
As dedicated rollout capacity grows, the benefit of toggle decreases, while PD disaggregation remains useful because it removes prefill--decode interference when computation is less constrained.
The gate may slightly reduce the measured gain in some situations, but it prevents toggle from being used when its estimated cost exceeds its benefit; in all tested configurations, the gated system remains faster than the baseline.
Together, PD disaggregation, intra-scale toggle, and the gate provide positive throughput gains across both models and all tested resource allocations.

\subsection{Workload Sensitivity}
\label{sec:eval-workload}

\begin{figure}[!t]
\centering
\includegraphics[width=0.96\columnwidth]{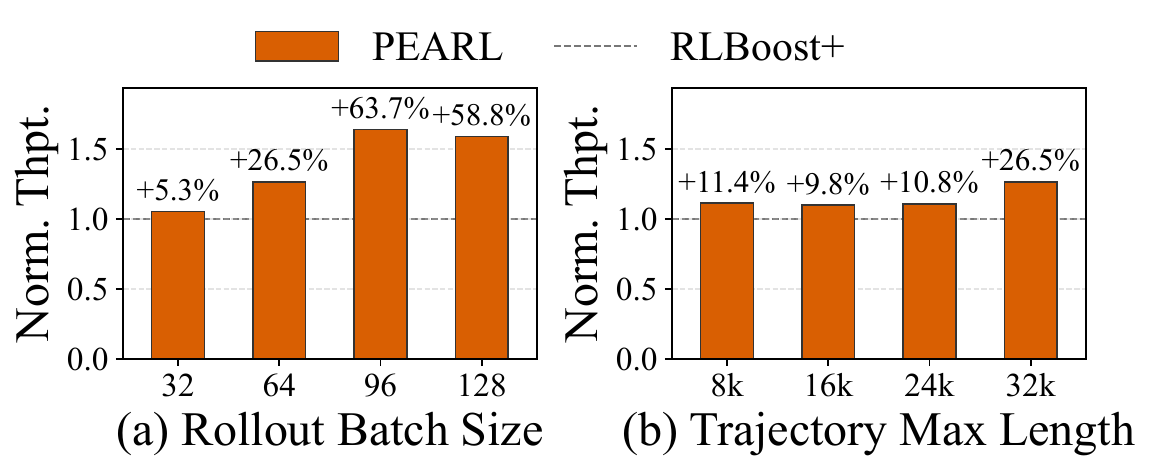}
\caption{Workload sensitivity. (a) Rollout batch size varies with the maximum trajectory length fixed at 32K tokens. (b) Maximum trajectory length varies with the rollout batch size fixed at 64.}
\label{fig:eval-impact-workload}
\end{figure}

We evaluate how rollout batch size and maximum trajectory length affect the throughput advantage of \sys{} over RLBoost+.
Both experiments use Qwen3-8B with 16 training GPUs and 8 dedicated rollout GPUs, varying one workload parameter at a time.
Figure~\ref{fig:eval-impact-workload} reports the results.

\noindent\textbf{Rollout batch size.}
We fix the maximum trajectory length at 32K tokens and vary the per-step rollout batch size from 32 to 128.
RLBoost+ reaches its highest measured throughput at batch size 64, whereas \sys{} peaks at 96; throughput declines beyond these respective peaks.
\sys{} improves throughput over RLBoost+ by 5.3--63.7\% across the tested batch sizes.
This pattern is consistent with the ablation results: intra-scale toggle provides more benefit when the dedicated rollout pool faces higher compute pressure, but less benefit when the workload offers limited opportunity to use additional capacity.

\noindent\textbf{Maximum trajectory length.}
We fix the per-step rollout batch size at 64 and vary the maximum trajectory length from 8K to 32K tokens.
Both systems' throughput decreases as the length limit increases, but \sys{} remains faster at every tested limit, with gains of 9.8--26.5\% over RLBoost+.

Together, the experiments show that \sys{} improves throughput across the tested Qwen3-8B workloads under a fixed resource budget, while the magnitude of the gain depends on the batch size and trajectory-length limit.

\subsection{Adaptation Quality and Cost}
\label{sec:eval-adaptation}

We evaluate the quality of the PD Decision Engine's ratio selection and the execution costs of inter-scale and intra-scale reconfiguration.

\subsubsection{PD Ratio Selection\texorpdfstring{\nopunct}{}}
\label{sec:eval-pd-accuracy}
\leavevmode\par
\vspace{0.5\baselineskip}
\noindent
We evaluate ratio selection using constant-resource intervals from the Qwen3-8B end-to-end runs with 10, 12, and 16 dedicated rollout GPUs, both with and without intra-scale borrowing.
For each interval, we reproduce the resource allocation and replay the recorded generation counts and lengths to benchmark every feasible PD ratio.
The ratio with the highest measured average throughput serves as the empirical best.

Table~\ref{tab:eval-pd-accuracy} shows that the selected ratios match the empirical best at all three dedicated-only resource points and at two of the three points with overlap enabled.
The only mismatch occurs with 10 dedicated GPUs and overlap enabled, where the selected $4P9D$ and the empirical best $5P8D$ differ by one worker's role.
In this case, throughput measured in the system run is approximately $0.4\%$ below the empirical optimum, although the comparison between separate runs may also reflect measurement variation.

\begin{table}[t]
    \centering
    \small
    \caption{PD-ratio selection for Qwen3-8B.}
    \label{tab:eval-pd-accuracy}
    \begin{tabular}{c c c c}
        \hline
        \textbf{Dedicated GPUs} & \textbf{Overlap} & \textbf{Selected} & \textbf{Empirical best} \\
        \hline
        10 & No  & $2P3D$   & $2P3D$   \\
        10 & Yes & $\boldsymbol{4P9D}$ & $\boldsymbol{5P8D}$ \\
        12 & No  & $3P3D$   & $3P3D$   \\
        12 & Yes & $4P10D$  & $4P10D$  \\
        16 & No  & $3P5D$   & $3P5D$   \\
        16 & Yes & $4P12D$  & $4P12D$  \\
        \hline
    \end{tabular}
\end{table}

\Needspace{5\baselineskip}
\subsubsection{Reconfiguration Cost\texorpdfstring{\nopunct}{}}
\label{sec:eval-overhead}
\leavevmode\par
\vspace{0.5\baselineskip}
\noindent
Table~\ref{tab:eval-reconfig} reports mean operation latencies for inter-scale expansion and intra-scale handoffs.
To interpret these costs, we distinguish their effects on the availability of additional rollout capacity, the admission of new requests, and the return of borrowed GPUs to training.

\begin{table}[t]
    \centering
    \small
    \caption{Mean reconfiguration latency in seconds.}
    \label{tab:eval-reconfig}
    \begin{tabular}{l r r}
        \hline
        \textbf{Operation} & \textbf{Qwen3-8B} & \textbf{Qwen3-30B-A3B} \\
        \hline
        Inter-scale preparation & 56.29 & 67.06 \\
        Inter-scale finalization & 11.05 & 17.89 \\
        \hline
        Intra-scale activation & 6.26 & 6.09 \\
        Intra-scale deactivation & 3.18 & 4.38 \\
        \hline
    \end{tabular}
\end{table}

\para{Inter-scale expansion.}
Inter-scale expansion first initializes new workers in a background preparation stage, which takes 56--67\,s across the two models.
Because existing workers continue rollout throughout preparation, this cost delays the availability of additional capacity without pausing the active rollout pool.
The subsequent finalization stage takes 11--18\,s to synchronize weights and update routing and communication groups.
During finalization, admission of new requests is paused while requests already in progress continue executing.

\para{Intra-scale activation and deactivation.}
Intra-scale reconfiguration avoids repeated worker initialization by retaining the training and rollout processes across handoffs and restoring or releasing their device state as needed.
Activation takes approximately 6\,s for both models and must complete before borrowed GPUs can contribute to rollout.
Deactivation takes 3.2--4.4\,s and must finish before these GPUs can resume training; dedicated rollout workers continue serving throughout both operations.
Because these costs recur with each borrowing cycle, the benefit-aware gate enables borrowing only when the predicted rollout-time saving exceeds the combined activation and return costs (\S\ref{sec:design-intra-scale}).
The fixed-resource ablations (\S\ref{sec:eval-ablation}) show that borrowing yields its largest throughput gains under limited dedicated rollout capacity, with smaller gains as dedicated capacity grows.

\para{Scale-in recovery.}
Scale-in preserves each interrupted request's generated prefix and interaction progress so that rollout can resume without restarting the trajectory (\S\ref{sec:implementation}).
For requests interrupted on decode workers, prefix-cache reuse limits additional prefill work to the newly generated tokens; interruptions on prefill workers require an additional full prefill pass.
In our measurements, this additional full-prefill pass takes less than 1\% of the time required to generate a complete trajectory, even at the maximum trajectory length.
This measurement quantifies the extra prefill work needed to recover an interrupted request and does not include the other operations involved in scale-in.

% \para{Retained state.}
% After deactivation, rollout instances retain 2.44\,GB per GPU for Qwen3-8B and 4.74\,GB per GPU for Qwen3-30B-A3B to preserve runtime state.
% This state remains on training GPUs rather than dedicated rollout GPUs, so it affects training-side memory headroom but does not increase the memory capacity of the dedicated rollout pool.
% Together, the asynchronous expansion path, short intra-scale handoff, and partial-trajectory recovery keep adaptation overhead localized; the remaining handoff cost is explicitly accounted for by the intra-scale gate.

\section{Related Work}
\label{sec:related}

\begingroup
\setlength{\emergencystretch}{3em}

\para{Agentic RL training systems.}
Many systems accelerate LLM RL post-training through asynchronous execution and rollout optimizations~\cite{areal,laminar,RhymeRL}.
Agentic RL frameworks further support multi-turn interactions with external environments~\cite{roll,rollart}.
RollArt, NeMo RL, and Slime also support PD-disaggregated rollout~\cite{rollart,nemorl_pd,slime}.
Their execution optimizations are complementary to deployment planning, which determines how the available GPUs should be used for rollout.
We implement \sys{} on ROLL and extend its rollout deployment to changing resource availability.
\sys{} selects colocated or disaggregated execution and the P/D ratio according to the workload and available GPUs, while keeping the training allocation fixed.

\para{RL with resource adaptation.}
RL systems exploit two forms of resource elasticity.
RLBoost~\cite{rlboost} harvests external preemptible GPUs, while ROSE~\cite{gao2026roserolloutservinggpus} shares spare serving capacity subject to serving SLOs; both provide \texttt{inter-scale} elasticity.
RLBoost temporarily reuse training GPUs during a seeding time window before training begins. However, this mechanism does not support asynchronous execution. More importantly, in agentic RL workloads, rollout latency is substantially higher than training latency. Consequently, even with a seeding time window, the training GPUs must still wait for rollouts after training starts, resulting in substantial idle bubbles.
SeamlessFlow~\cite{seamlessflow} and BiDiRL~\cite{bidirl} exploit \texttt{intra-scale} elasticity by reusing training resources for rollout.
BiDiRL accounts for predicted execution gains and switching costs when borrowing resources.
These approaches provide different sources of rollout capacity.
\sys{} considers both forms of elasticity together with PD execution planning, adapting the worker set and its role allocation while accounting for the cost of transitions.

\para{LLM serving techniques for RL rollouts.}
Splitwise~\cite{splitwise} supports separated and mixed PD execution, while DOPD~\cite{dopd} adjusts instance counts and the P/D ratio in response to serving load.
For multi-turn inference, AMPD~\cite{ampd} combines online prefill routing with offline resource planning, whereas PPD~\cite{ppd} routes append-prefill between prefill and decode instances based on latency trade-offs.
These systems optimize serving latency, throughput, or goodput.
\sys{} instead selects the execution mode and P/D ratio to reduce rollout-batch makespan as external resources scale and training GPUs are borrowed or returned.
\par\endgroup

\section{Conclusion}
\label{sec:conclusion}

% We present \sys{}, an asynchronous agentic RL system that jointly adapts preemptible-GPU usage, idle training-GPU reuse, and PD execution modes and ratios. A unified orchestration layer applies these decisions with low overhead. On SWE-bench under preemptible-GPU traces, \sys{} achieves $2.17$--$2.79\times$ the end-to-end throughput of fixed-resource ROLL and improves throughput over RLBoost+ by up to 26.9\% for Qwen3-8B and 35.8\% for Qwen3-30B-A3B.

We present \sys{}, an asynchronous agentic RL system for dynamic GPU resources that jointly adapts rollout resources, idle training-GPU reuse, and PD execution roles. It minimizes batch makespan and applies decisions through a unified orchestration layer: the Inter-Scale Cluster Rebuilder handles preemptible-GPU expansion with lazy weight propagation and atomic admission, while the Intra-Scale Toggle Manager borrows idle training GPUs via process retention and device-state handoff. 
On SWE-bench under preemptible-GPU traces, \sys{} achieves $2.17$--$2.79\times$ the end-to-end throughput of fixed-resource ROLL and improves throughput over RLBoost+ by up to 26.9\% and 36.3\% for Qwen3-8B and Qwen3-30B-A3B, respectively.
% Evaluation on SWE-bench under preemptible-GPU traces shows that \sys{} achieves $2.17$--$2.79\times$ the end-to-end throughput of fixed-resource ROLL across the two models. Compared with RLBoost+, throughput improves by up to approximately 26.9\% for Qwen3-8B and 35.8\% for Qwen3-30B-A3B, demonstrating the benefit of jointly adapting resources and roles for elastic agentic RL.

\clearpage

\bibliographystyle{ACM-Reference-Format}
\bibliography{sample_base}

\end{document}